\documentclass[journal]{IEEEtran}
\usepackage{amsmath,amsfonts}
\usepackage{algorithmic}
\usepackage{algorithm}
\usepackage{array}
\usepackage[caption=false,font=normalsize,labelfont=sf,textfont=sf]{subfig}
\usepackage{textcomp}
\usepackage{stfloats}
\usepackage{url}
\usepackage{verbatim}
\usepackage{graphicx}
\usepackage{cite}
\usepackage{float}
\usepackage{hyperref}
\usepackage{amsmath}
\usepackage{amssymb}
\usepackage{booktabs}   % 美化表格横线
\usepackage{multirow}   % 多行合并
\usepackage{makecell}   % 允许表格单元格换行
\usepackage{adjustbox}  % 另一种表格缩放方案（可选）
\usepackage{rotating}   % 旋转表格文本
\usepackage{colortbl}
\usepackage[table]{xcolor}
\usepackage{bbding}  
\definecolor{iccvblue}{rgb}{0.21,0.49,0.74}

\definecolor{Acolor}{RGB}{172,114,185}
\definecolor{Bcolor}{RGB}{225, 157, 117}
\definecolor{Ccolor}{RGB}{255, 185, 76}
\definecolor{Dcolor}{RGB}{146, 208, 116}

\begin{document}

\title{IAP: Improving Continual Learning of Vision-Language Models via Instance-Aware Prompting}

\author{Hao Fu, Hanbin Zhao$^{\dag}$, Jiahua Dong, Henghui Ding, Chao Zhang and Hui Qian
        % <-this % stops a space
\thanks{Hao Fu, Hanbin Zhao, Chao Zhang and Hui Qian are with Zhejiang University, Hangzhou 310027, China (e-mail: haof.pizazz@zju.edu.cn; zhaohanbin@zju.edu.cn; zczju@zju.edu.cn; qianhui@zju.edu.cn).}% <-this % stops a space
\thanks{Jiahua Dong is with Mohamed bin Zayed University of Artifical Intelligence, Abu, Dhabi, United Arab Emirates. (e-mail: dongjiahua1995@gmail.com).}
\thanks{Henghui Ding is with Fudan University, Shanghai, China 200433. (e-mail: henghui.ding@gmail.com).}
\thanks{The corresponding author is Hanbin Zhao.}
}

\markboth{IEEE TRANSACTIONS ON IMAGE PROCESSING}%
{Shell \MakeLowercase{\textit{et al.}}: A Sample Article Using IEEEtran.cls for IEEE Journals}


\maketitle

\begin{abstract}
Recent pre-trained vision-language models (PT-VLMs) often face a Multi-Domain Task Incremental Learning (MTIL) scenario in practice, where several classes and domains of multi-modal tasks are incrementally arrived. Without access to previously seen tasks and unseen tasks, memory-constrained MTIL suffers from forward and backward forgetting. To alleviate the above challenges, parameter-efficient fine-tuning techniques (PEFT), such as prompt tuning, are employed to adapt the PT-VLM to the diverse incrementally learned tasks. To achieve effective new task adaptation, existing methods only consider the effect of PEFT strategy selection, but neglect the influence of PEFT parameter setting (e.g., prompting). In this paper, we tackle the challenge of optimizing prompt designs for diverse tasks in MTIL and propose an Instance-Aware Prompting (IAP) framework. Specifically, our Instance-Aware Gated Prompting (IA-GP) strategy enhances adaptation to new tasks while mitigating forgetting by adaptively assigning prompts across transformer layers at the instance level. Our Instance-Aware Class-Distribution-Driven Prompting (IA-CDDP) improves the task adaptation process by determining an accurate task-label-related confidence score for each instance. Experimental evaluations across 11 datasets, using three performance metrics, demonstrate the effectiveness of our proposed method. The source codes are available at \url{https://github.com/FerdinandZJU/IAP}.
\end{abstract}


\section{Introduction}
\begin{figure}[t]
  \centering
  \includegraphics[width=0.48\textwidth]{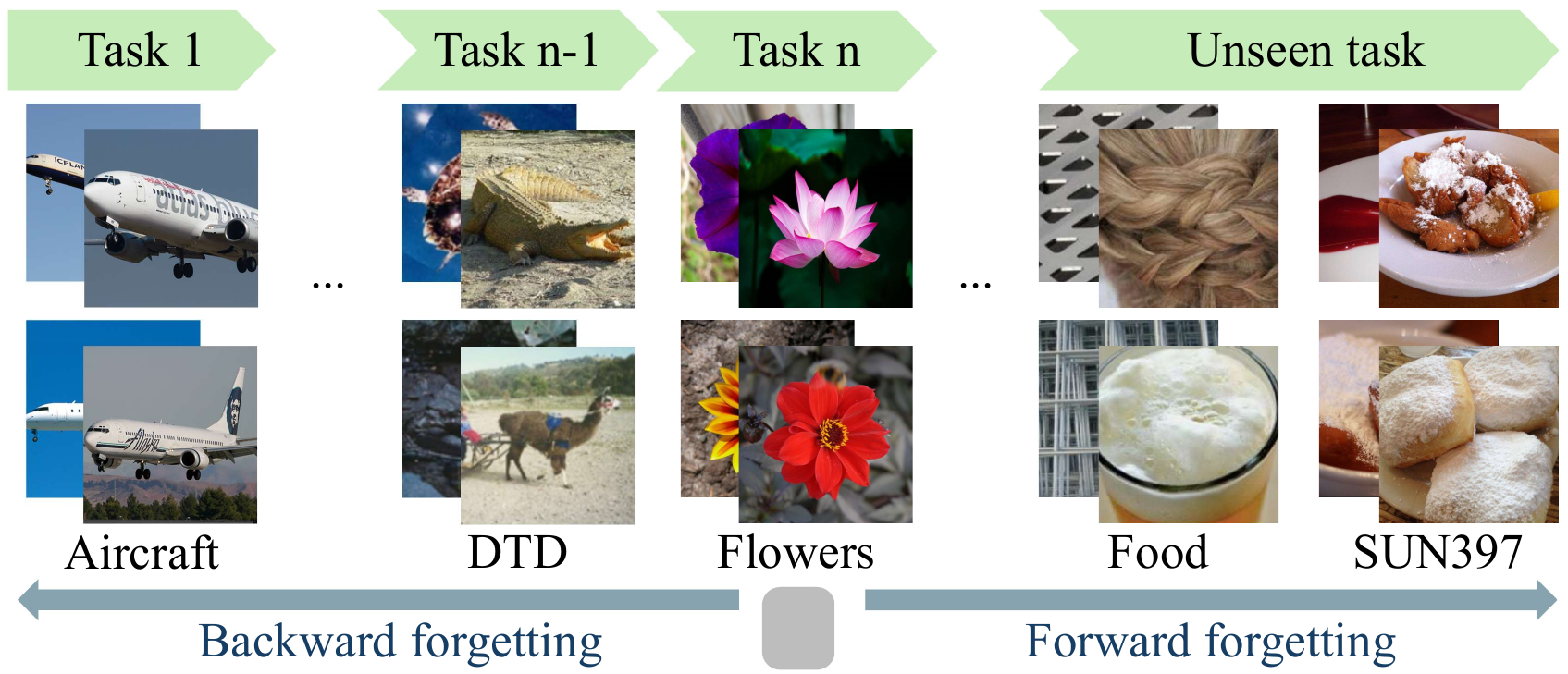}
  \caption{Illustration of backward forgetting and forward forgetting. During the current learning process (e.g., Flowers), backward forgetting refers to the degradation of seen tasks (e.g., Aircraft and DTD); forward forgetting refers to the decline in zero-shot generalization capability on unseen tasks (e.g., Food and SUN397).}
  \label{fig:intro}
\end{figure}

\IEEEPARstart{R}{ecent} years have witnessed a great development of deep neural networks in numerous multi-modal applications, where all the required data are available simultaneously for training on various tasks~\cite{zhao2024learning, zhou2024continual}. Nevertheless, real-world applications usually meet a challenging Multi-Domain Task Incremental Learning (MTIL) scenario~\cite{zheng2023preventing}: 1) tasks of multi-modal data are acquired incrementally, 2) the task distribution and target classes vary across tasks. Since the learned tasks data are un-available in a memory-constrained MTIL, multi-modal models suffer from the catastrophic forgetting phenomenon~\cite{zhou2024continual} from two aspects: backward forgetting and forward forgetting~\cite{zheng2023preventing}, as shown in Figure \ref{fig:intro}.

Due to the remarkable zero-shot ability of vision-language models (VLMs)~\cite{floridi2020gpt, xuan2024adapting, lu2025decoupling, liu2025museummaker} (e.g., CLIP~\cite{radford2021learning}), existing works have explored the multi-modal class incremental learning with pre-trained VLMs and parameter-efficient fine-tuning (PEFT) techniques (e.g., prompt, adapter, LoRA)~\cite{zhou2024continual, zhao2024learning}. Typically, these pre-trained model-based multi-modal class incremental learning works usually maintain a fixed pre-trained VLM and incrementally learn few task-specific parameters to adapt to various encountered tasks. Since the memory of MTIL is constrained, such pre-trained model-based methods mainly focus on designing an appropriate PEFT strategy for effective adaptation on various tasks. The selection of PEFT strategy—whether prompt, adapter, or LoRA—as well as the configuration of PEFT parameters, including the placement and quantity of prompts, adapters, or LoRA modules, significantly influences the performance of MTIL for acquiring new tasks.

Existing MTIL works have widely discussed the effect of PEFT strategy selection on MTIL performance, but overlook the analysis of the PEFT parameter settings. These works typically employ a manually designed PEFT parameter setting and apply them consistently across the incremental learning of different tasks. For instance, when learning a new task, prompt-based MTIL methods (e.g., DIKI~\cite{tang2024mind}) consistently use the same number of prompts at the same positions within the pre-trained VLM (an operation we refer to as \emph{prompting}). However, our observations suggest that the optimal prompting strategy for adapting to different tasks varies within the MTIL framework. Thus, the key challenge lies in developing a flexible and adaptive prompting strategy to enhance task adaptation.

Motivated by the above observations, we propose a novel Instance-Aware Prompting (IAP) framework for MTIL, which adaptively allocates prompt positions and weights at the instance level. Specifically, we introduce an Instance-Aware Gated Prompting (IA-GP) strategy that determines whether to apply prompts in each transformer layer based on instance-specific features. Furthermore, we propose an Instance-Aware Class-Distribution-Driven Prompting (IA-CDDP) strategy to compute more reliable confidence scores at the instance level, which are then used as prompt weights in subsequent operations. By integrating above two strategies, we solve the problems of fixed prompting strategy of conducting incremental learning on pre-trained VLMs.

The contributions of our proposed IAP approach can be summarized in threefold:  
\begin{itemize}
    \item We design an Instance-Aware Gated Prompting strategy to address the challenge of determining where to prompt. We enhance PEFT techniques, enabling the model to dynamically allocate prompting positions at the instance level, thereby improving the models incremental learning capability.
    \item We introduce an Instance-Aware Class-Distribution-Driven Prompting strategy to derive more reliable prompt weights. To enhance model performance across diverse instances, we employ a two-stage distribution modeling strategy that operates at both the task and class levels during inference.
    \item Extensive experiments on benchmark datasets demonstrate that our method achieves state-of-the-art (SOTA) performance while utilizing only 1.18\% of the training parameters, outperforming all existing methods.
\end{itemize}

\section{Related Works}
\subsection{Incremental Learning}

Conventional incremental learning can be categorized into three types: task-incremental learning (TIL)~\cite{hsu2018re,van2019three,zheng2023preventing, fan2017hd}, domain-incremental learning (DIL)~\cite{zhao2021and}, and class-incremental learning (CIL)~\cite{zhao2021memory,feng2024lw2g, wang2025class, liu2024ntk, ji2023memorizing,zhao2024sdcot++,wang2024model}. Among these, task-incremental learning requires explicit task identifies during inference, while class-incremental learning requires non-overlapping classes across different tasks.
Three technologies are generally utilized in incremental learning: 1) Regularization-based methods, such as Elastic Weight Consolidation (EWC)~\cite{kirkpatrick2017overcoming} and Memory Aware Synapses (MAS)~\cite{aljundi2018memory}. These methods introduce regularization terms to constrain optimization directions and mitigate catastrophic forgetting. 2) Rehearsal-based methods~\cite{wu2019large,hou2018lifelong,hou2019learning,park2021class, zhou2024balanced}, Representative methods are Learning without Forgetting (LwF)~\cite{li2017learning} and Incremental Classifier and Representation Learning (iCaRL)~\cite{rebuffi2017icarl}. These approaches require additional storage to retain either previous model parameters or representative samples from seen tasks. 3) Network expansion-based methods~\cite{xu2018reinforced,li2019learn}, with Dynamic Expandable Networks (DEN)~\cite{yoon2017lifelong} being a representative approach. These methods dynamically expand the neural network structure to accommodate new tasks. Recently, observing the strong generalization of pre-trained models, some incremental learning approaches are designed to fine-tune the pre-trained models~\cite{jung2023generating,tang2023prompt,zhou2025learning,chen2024promptfusion,gao2023unified}, which called parameter-efficient fine-tuning (PEFT). PEFT techniques achieve adaptation for downstream tasks by fine-tuning only a small number of model parameters or maintaining a limited set of additional parameters while keeping the majority of the pre-trained model's parameters frozen. PEFT technology significantly reduces computational and storage costs and mitigates the catastrophic forgetting. Notable related methods include L2P~\cite{wang2022learning}, DualPrompt~\cite{wang2022dualprompt}, S-Prompt~\cite{wang2022s}, and CODA-Prompt~\cite{smith2023coda}. However, these methods only consider stable distributions and lack of the capability to learn tasks with distinct distributions. Our work focus on MTIL benchmark, where data distributions are distinct and arrive continually. 

\subsection{Downstream tasks of vision-language models} 
Visual Language Models (VLMs)~\cite{jia2021scaling,pham2023combined} have achieved remarkable progress in multi-modal research, successfully enabling cross-modal understanding and generation capabilities through joint modeling of visual inputs and natural language. With the explosive growth of online data, recent years have witnessed further advancements in VLMs driven by larger-scale models and more extensive datasets. However, when researchers attempt to fine-tune the VLMs for a downstream task, an inevitable degradation occurs in their zero-shot capabilities of other tasks~\cite{kumar2022fine,chen2023forgetting}. Various approaches have been proposed to mitigate the degradation. ZSCL~\cite{zheng2023preventing} leverages the knowledge distillation technology~\cite{li2017learning}, treating the original VLM as a teacher model and distilling knowledge into the fine-tuned model through constructed ``wild''~\cite{chen2021learning} datasets (e.g., ImageNet~\cite{deng2009imagenet}). Although ZSCL partially alleviates forgetting of pre-trained knowledge, it requires additional storage space for ``wild'' datasets, and its full-parameter fine-tuning strategy incurs substantial computational costs. Alternative approaches such as MoE-Adapter~\cite{yu2024boosting} and DIKI~\cite{tang2024mind} employ PEFT technology, updating only a small set of parameters to reduce the forgetting of pre-trained knowledge. Nevertheless, these methods continue to encounter challenges associated with computational resource overhead and the under-exploration of instance-aware features. Our proposed approach overcomes these limitations by introducing an instance-aware prompting strategy, significantly enhancing model performance.

\section{Methodology}
\subsection{Preliminaries}
\subsubsection{MTIL Benchmark} 
Given a pre-trained VLM, which incrementally learns from a stream of tasks originating from $\mathcal{T}$ distinct domains ${\{D_{1}, D_{2},...,D_{\mathcal{T}}\}}$, each domain consists of ${N}$ samples, denoted as $D_{t}=(x^{t}_{n},y^{t}_{n})^{N}_{n=1}$, where ${x_n}$ is an image and $y_n$ is the corresponding one-hot ground truth. For domain $D_{n}$, the associated class set is given by $C_n=\{c^n_i\}^{M^n}_{i=1}$, where each $c^n_{i}$ represents a textual label. In the MTIL setting, the domain ${D_n}$ is accessible only during its corresponding $n$-th incremental learning session. Moreover, class sets are disjoint across domains, i.e., $C_i \neq C_j$ for any $i \neq j$, and the data distributions also distinct across domains, meaning $\mathbb{P}_i \neq \mathbb{P}_j$, where $\mathbb{P}$ denotes the distribution. During the inference, the model is evaluated without accessing to the task identifier, requiring it to infer across all previously seen domains without explicit domain information.

\subsubsection{CLIP Model} The pre-trained VLMs such as CLIP~\cite{radford2021learning} consisted of an image encoder $F_v$ and a text encoder $F_t$. For a class $c_i^n$, CLIP model first transforms it to a sentence by a template such as ``\{a photo of \{$c_i^n$\}\}'', and then encodes it into text embedding $t_i$. CLIP model is trained by leverages contrastive loss, the objective can be defined as:
\begin{equation}
    L = - \sum_{i=1}^{N} \log \left( \frac{ \exp \left( \text{sim} \left( F_{v}(x_i), F_{t}(t_i) \right) / \tau \right)}{ \sum_{j=1}^{N} \exp \left( \text{sim} \left( F_{v}(x_i), F_{t}(t_j) \right) / \tau \right)} \right),
\end{equation}
where $\tau$ is the temperature, $\text{sim}(u, v) = \frac{u^T \cdot v}{\|u\| \|v\|} $ is the cosine similarity function, the contrastive loss enables the CLIP model to capture the inter-modal similarity between image embedding and all text embeddings.

\subsubsection{Interference-Free Knowledge Integration (IKI) Mechanism} We follow ~\cite{tang2024mind} to leverage a prompt-based incremental learning method. Specifically, a set of prompt pools are maintained for a stream of tasks, which can be denoted as $\mathcal{P} = \{ \mathcal{P}_1, \mathcal{P}_2, \dots, \mathcal{P}_\mathcal{T} \}$, where $ P_t = (K_t, V_t)$, and $K_t,V_t \in \mathbb{R}^{l \times d}$. $l$ is the prompt length, $d$ is the embedding dimension. When a test sample $x_n$ comes, IKI mechanism first selects the corresponding prompt $\mathcal{P}_r$, and produces a task-specific attention output, which can be formulated as:
\begin{equation}
    O_{r} = \text{softmax} \left( \frac{Q_n K_r^T}{\sqrt{d}} \right) V_r,
\end{equation}
where $O_{r}, Q_n \in \mathbb{R}^{L\times d}$, and $Q_n$ is original query vector of CLIP model, $L$ is the length of embedded feature, $K_r$ and $V_r$ are come from the selected prompt $\mathcal{P}_r$. IKI mechanism leverages a residual branch for the top layers in a transformer architecture:
\begin{equation}
    O_{IKI} = O_{ori} + O_{r},
\end{equation}
where $O_{ori}$ is the original attention output of CLIP model. IKI mechanism incrementally learning different tasks by maintaining different prompt pools for each distribution.

\begin{figure*}[t]
  \centering
  \includegraphics[width=1.0\textwidth]{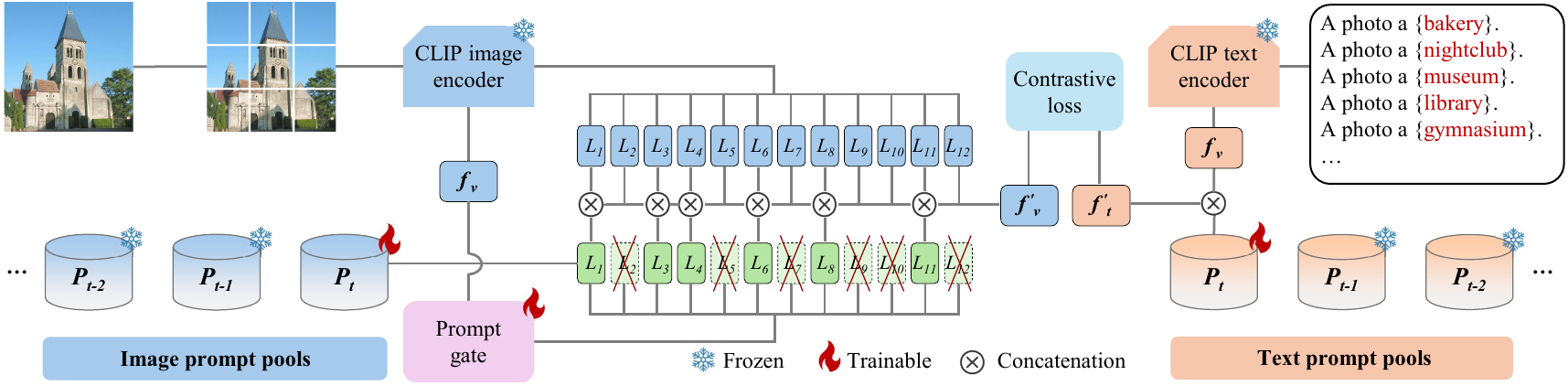}
  \caption{Illustration of Instance-Aware Gated Prompting strategy. The figure illustrates the processing workflow of the IA-GP strategy applied to an instance ``abbey''. The image is segmented into patches initially, which are then fed into the visual encoder of the CLIP model. The prompt pools of seen tasks are kept frozen. IA-GP leverages the visual features from the original CLIP encoders, denoted as $f_v=F_v(x)$, as input of prompt gate module. Hard Gumbel logits produced by prompt gate modules are used to determine whether to retrieve prompts from prompt pools. $L$ denoted each self-attention layer in the Transformer architectures. Outputs from the original CLIP model are represented in blue, while those incorporating retrieved prompts via the IA-GP strategy are shown in green. 
  The processed visual and textual features are optimized through a contrastive loss function.}
  \label{fig:gated_prompting}
\end{figure*}
\subsection{Instance-Aware Prompting}
\subsubsection{Gated Prompting}
In transformer architectures, the efficacy of a uniform prompting strategy across tasks with diverse distributions remains a critical challenge. For example, EuroSAT~\cite{helber2019eurosat} (10 classes) and SUN397~\cite{xiao2010sun} (397 classes) differ in distribution and granularity, and the differences become particularly significant when analyzing individual instances within a dataset, suggesting that prompting configurations must be adaptive to instance-specific characteristics. To address this, we propose Instance-Aware Gated Prompting (IA-GP), which employs an instance-aware gating mechanism to dynamically tailor the prompting strategy to the individual features.

The IA-GP strategy incorporates multiple prompting gates positioned before the vision transformer layers, as illustrated in Figure \ref{fig:gated_prompting}. Each prompt gate consists of a Gumbel linear function $H_i$, which maps the embedded image features to a $K$-dimensional space, and a Gumbel distribution which used to generate the samples uniformly. For each instance, we utilize the image features extracted by the original CLIP model as the input to the IA-GP strategy. This operation is motivated by the fact that the original CLIP model is frozen, ensuring that its raw image features reside in a stable distribution. We then compute the instance-aware Gumbel logit by:
\begin{equation}
    G_i = \frac{\exp (\log\left({H_i(F_v(x)) + g_i})/{\tau}\right)}{\sum_{j=1}^K \exp (\log\left({H_i(F_v(x)) + g_j})/{\tau}\right)}.
\end{equation}

In our approach, we set $K=2$ to implement a hard Gumbel Softmax mechanism, which facilitates binary decision-making for prompting. $g_i = -\log(-\log(U_i))$ represents a random logit sampled from the Gumbel distribution, where $U_i \sim U(0,1)$. The parameter $\tau$ denotes the temperature of the IA-GP strategy, controlling the sharpness of the decision boundary. The IA-GP strategy generates a two-dimensional output $G_i$, which is subsequently transformed into a one-hot vector via the hard Gumbel Softmax operation. This mechanism enables IAP approach to dynamically determine whether to prompt for each layer at the instance level. Specifically, if prompting is beneficial, the IA-GP strategy adjusts the Gumbel logit such that $G_i$ approaches 1. Conversely, if prompting operation is detrimental, the gate sets $G_i$ close to 0, thereby preserving the original output of the CLIP model. In summary, for each transformer layer during training, the output of our IA-GP strategy is formulated as follows:
\begin{equation}
O_{IAP} = O_{ori} + \text{one-hot}(G) O_{r}.
\end{equation}
The operations described above are implemented for each individual instance. As a result, the IA-GP strategy can dynamically determine and assign the appropriate prompting layers according to the specific features of each instance.
\subsubsection{Class-Distribution Driven Prompting}
Incremental VLMs must mitigate backward forgetting (degradation of learned distributions) and prevent forward forgetting (impaired generalization to unseen distributions) to preserve zero-shot generalization. In the MTIL scenario, where task identifiers are absent during inference, distinguishing seen tasks from unseen tasks is crucial. DIKI~\cite{tang2024mind} adjusts prompt weights using instance-to-task similarity as a confidence score. However, this approach is suboptimal, where high-confidence instances should use unadjusted prompts directly, while low-confidence instances risk noise if prompted, the original CLIP model needs to be employed. For instances with intermediate confidence scores, the confidence score needs to be more reliable. To address this, we propose Instance-Aware Class-Distribution-Driven Prompting (IA-CDDP), a two-stage strategy that reassesses instances via both task- and class-specific perspectives during inference.

\begin{figure*}
  \centering
  \includegraphics[width=1\textwidth]{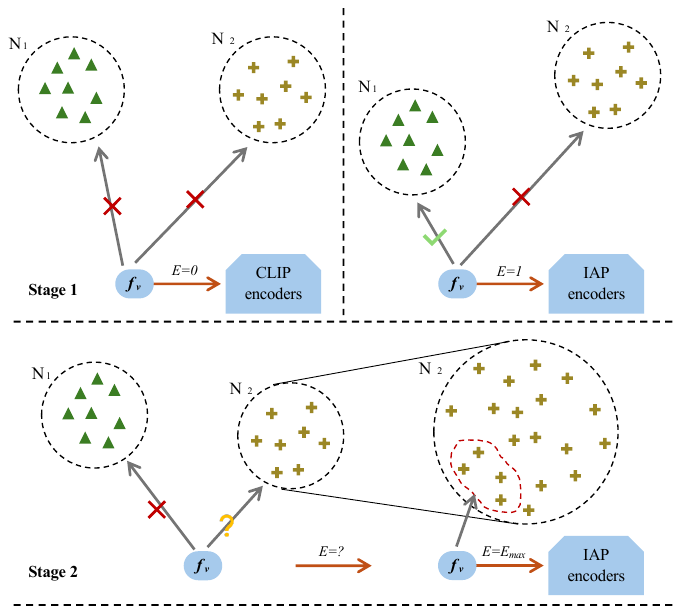}
  \caption{Illustration of IA-CDDP strategy. The IA-CDDP strategy leverages visual features from the original frozen CLIP model for a two-stage confidence assessment per instance: 1) confidence scores between the instance and seen distributions are evaluated and binarized using predefined thresholds.2) for median confidence, the top $K$ highest scores are selected by modeling the in-task class distribution, with their mean computed as the final prompting weight.
  }
  \label{fig:class-driven}
\end{figure*}
During the training phase, for each newly encountered task with the distribution $D_i$, the visual features of the images within $D_i$ are extracted using the original frozen CLIP image encoder. Subsequently, the mean vector $\mu_i$ and covariance matrix $\Sigma_i$ of these feature vectors are computed and stored:
\begin{equation}
\begin{aligned}
    &\mu_i = \mathbb{E}_{x_i \sim D_i} \left[ F_v(x_i) \right] \\
    &\Sigma_i = \mathbb{E}_{x_i \sim D_i} \left[ \left( F_v(x_i) - \mu_i \right)^T \left( F_v(x_i) - \mu_i \right) \right],
\end{aligned}
\end{equation}
where $F_v(x_i)$ represents the feature vector extracted for each image $x_i \in D_i$ by the frozen CLIP image encoder, ensuring that the features reside in a stable feature space due to the fixed encoder parameters. In the inference stage, for each learned distribution $D_i$, a multivariate Gaussian distribution $\mathcal{N}_i(\mu_i, \Sigma_i)$ is constructed using the stored $\mu_i$ and $\Sigma_i$. For a given test sample, its feature vector is extracted using the same CLIP image encoder, and the log-likelihood of this feature vector under each $\mathcal{N}_i$ is calculated to assess its fit to the learned distributions:
\begin{equation}
\begin{aligned}
E_i'&= \log \varphi(F_v(\hat{x}); \mu_i, \Sigma_i) \\
    &= -\frac{1}{2} \left[ (F_v(\hat{x}) - \mu_i)^T \left( \Sigma_i \right)^{-1} (F_v(\hat{x}) - \mu_i) \right.  \\
    & + d \log 2\pi + \log |\Sigma_i| \Big],
\end{aligned}
\end{equation}
% \begin{equation}
% E_i'= \log \varphi(F_v(\hat{x}); \mu_i, \Sigma_i) = -\frac{1}{2} \left[ (F_v(\hat{x}) - \mu_i)^T \left( \Sigma_i \right)^{-1} (F_v(\hat{x}) - \mu_i) \right. + d \log 2\pi + \log |\Sigma_i| \Big],
% \end{equation}
$\hat{x} \in \mathbb{R}^d$ denote the image input of a test sample, where $d$ represents the dimensionality of the image features. and $\varphi$ denote the probability density function (PDF) corresponding to the learned distribution associated with task $i$. The IKI mechanism computes the PDF values $E'_i = \varphi(\hat{x})$ for each task $i$ and determines the maximum score as $E'_{max}=\text{max}_{i\in[1,\mathcal{T}]}E'_i$. Subsequently, it applies the sigmoid function $\sigma$ to obtain the confidence score, yielding $E_{max}=\sigma E'_{max}$, which maps $E'_{max}$ to the interval [0,1]. This value $E_{max}$ is then employed as the prompting weight within the attention layers of both the vision transformer and the text transformer. The IA-CDDP we proposed is designed with a two-stage architecture, as shown in Figure \ref{fig:distributions}. In the first stage, the module evaluates the confidence scores of the test sample $\hat{x}$ across the $\mathcal{T}$ tasks. For samples exhibiting significantly high or low confidence scores, the module does not assign a weight to the prompt. Instead, it directly utilizes either the incremental learning model or the original CLIP model, depending on the specific bounds:
\begin{equation}
\hat{E}_{max} = 
\begin{cases}
    0 & \text{for } E_{max} \leq \text{lower bound} \\
    1 & \text{for } E_{max} \geq \text{upper bound} \\
    E_{max} & \text{into the second stage}
\end{cases}
\quad\quad,
\end{equation}
for samples that yield confidence scores within a median range, we employ the second stage of our IA-CDDP approach, which reconstructs the Gaussian distribution for each class within the distribution $D_i$:
\begin{equation}
\begin{aligned}
    E'_{i,j} &= \log \varphi(F_v(\hat{x}); \mu_{i,j}, \Sigma_{i,j}) \\
    &= -\frac{1}{2} \left[ (f(\hat{x}) - \mu_{i,j})^T \left( \Sigma_{i,j} \right)^{-1} (f(\hat{x}) - \mu_{i,j}) \right. \\
    & + d \log 2\pi + \log |\Sigma_{i,j}| \Big],
\end{aligned}
\end{equation}
where $\mu_{i,j}$ and $\Sigma_{i,j}$ denote the mean and covariance matrix of class $j$ within distribution $D_i$. For every specific instance, we then compute the mean of the confidence scores associated with the most similar $K$ classes. The mean of these $K$ classes serves as the final weight for the prompt, which is subsequently integrated into the attention layers of the transformer model:
\begin{equation}
\hat{E}_{means} = \frac{1}{K}\sum_{k=1}^{K}\sigma E'_{[i]},
\end{equation}
where $E_{[i]}$ represent the top $K$ confidence scores corresponding to the most similar classes within the distribution. Through the second stage of the IA-CDDP strategy, test samples that are distant from the Gaussian distribution of the entire task can still obtain a reasonable confidence score.

\begin{table*}[t]
    \centering
    \caption{Comparison with SOTA on MDCIL benchmark in terms of ``Transfer'', ``Average'', and ``Last'' metrics (\%). ``Ours" denotes our method. We label the best and second methods with \textbf{bold} and \underline{underline} styles. The presented results are derived from the Order-$\text{I}$.}
    {\fontsize{9pt}{11pt}\selectfont
    \renewcommand{\arraystretch}{1.0}
    \resizebox{1.0\linewidth}{!}{
    \begin{tabular}{c>{\raggedright\arraybackslash}p{3cm} >{\centering\arraybackslash}p{1cm} >{\centering\arraybackslash}p{1cm}>{\centering\arraybackslash}p{1cm} >{\centering\arraybackslash}p{1cm} >{\centering\arraybackslash}p{1cm}>{\centering\arraybackslash}p{1cm} >{\centering\arraybackslash}p{1cm} >{\centering\arraybackslash}p{1cm}>{\centering\arraybackslash}p{1cm} >{\centering\arraybackslash}p{1cm} >{\centering\arraybackslash}p{1cm} >{\centering\arraybackslash}p{1.5cm}}
        \toprule
           & {\hspace{1em}} Method & \makecell[c]{\rotatebox{90}{Aircraft~\cite{maji2013fine}}} & \makecell[c]{\rotatebox{90}{Caltech101~\cite{fei2004learning}}} & \makecell[c]{\rotatebox{90}{CIFAR100~\cite{krizhevsky2009learning}}} & \makecell[c]{\rotatebox{90}{DTD~\cite{cimpoi2014describing}}} & \makecell[c]{\rotatebox{90}{EuroSAT~\cite{helber2019eurosat}}} & \makecell[c]{\rotatebox{90}{Flowers~\cite{nilsback2008automated}}} & \makecell[c]{\rotatebox{90}{Food~\cite{bossard2014food}}} & \makecell[c]{\rotatebox{90}{MNIST~\cite{deng2012mnist}}} & \makecell[c]{\rotatebox{90}{OxfordPet~\cite{parkhi2012cats}}} & \makecell[c]{\rotatebox{90}{Cars~\cite{krause20133d}}} & \makecell[c]{\rotatebox{90}{SUN397~\cite{xiao2010sun}}} & \makecell[c]{{\textit{Average}}} \\
  
        \midrule
        
            \multirow{2}{*}{\rotatebox{90}{CLIP}}& {\hspace{1em}}Zero-shot & 24.3 & 88.4 & 68.2 & 44.6 & 54.9 & 71.0 & 88.5 & 59.4 & 89.0 & 64.7 & 65.2 & 65.3  \\
            & {\hspace{1em}}Full Fine-tune & 62.0 & 95.1 & 89.6 & 79.5 & 98.9 & 97.5 & 92.7 & 99.6 & 94.7 & 89.6 & 81.8 & 89.2  \\
            \midrule\midrule
            \multirow{12}{*}{\rotatebox{90}{\textbf{Transfer}}} &{\hspace{1em}}Continual-FT & & 67.1 & 46.0 & 32.1 & 35.6 & 35.0 & 57.7 & 44.1 & 60.8 & 20.5 & 46.6 & 44.6 \\
            & {\hspace{1em}}LwF \cite{li2017learning} &  & 74.5 & 56.9 & 39.1 & \textbf{51.1} & 52.6 & 72.8 & {60.6} & 75.1 & 30.3 & 55.9 & 58.9 \\
            & {\hspace{1em}}iCaRL \cite{rebuffi2017icarl} &  & 56.6 & 44.6 & 32.7 & 39.3 & 46.6 & 68.0 & 46.0 & 77.4 & 31.9 & 60.5 & 50.4 \\
            & {\hspace{1em}}LwF-VR \cite{ding2022don} & & 77.1 & 61.0 & 40.5 & 45.3 & 54.4 & 74.6 & 47.9 & 76.7 & 36.3 & 58.6 & 57.2  \\
            & {\hspace{1em}}WiSE-FT \cite{wortsman2022robust} & & 73.5 & 55.6 & 35.6 & 41.5 & 47.0 & 68.3 & 53.9 & 69.3 & 26.8 & 51.9 & 52.3  \\
            & {\hspace{1em}}ZSCL \cite{zheng2023preventing} & & 86.0 & 67.4 & \textbf{45.4} & \underline{50.4} & {69.1} & \underline{87.6} & {61.8} & 86.8 & 60.1 & \textbf{66.8} & {68.1} \\
            % & {\hspace{1em}}AweForget \cite{zheng2023preventing} & & 88.8 & 68.4 & \textbf{46.1} & 56.2 & \textbf{70.6} & \textbf{87.9} & 62.4 & 88.1 & 62.2 & \textbf{67.0} &  \\
            % \rowcolor{Thistle!20}

            \cmidrule{2-14}
            & {\hspace{1em}}L2P \cite{wang2022learning} & & 65.6 & 50.9 & 30.4& 41.4 & 49.3 & 71.8 & 36.3 & 77.5 & 55.3 & 53.4& 53.2 \\
            & {\hspace{1em}}DualPrompt \cite{wang2022dualprompt} & & 56.7 & 51.4& 28.7& 33.7 & 45.6 & 70.9 & 59.5& 77.7 & 49.5 & 50.4 & 52.4\\
            & {\hspace{1em}}S-Prompts \cite{wang2022s} & & 67.3 & 49.4 & 26.7& 39.7 & 47.1 & 70.2 & 34.3 & 78.9 & 56.7 & 52.2 & 52.2\\
            & {\hspace{1em}}MoE-Adapter \cite{yu2024boosting}& & {87.9} & {68.2} & \underline{44.4} & 49.9 & \textbf{70.7} & \textbf{88.7}& 59.7 & \underline{89.1} & \underline{64.5} & 65.5 & \underline{68.9}\\
            & {\hspace{1em}}DIKI \cite{tang2024mind} & & \underline{92.9} & \textbf{69.1} & 43.2 & 43.9 & 65.4 & {85.3} & \underline{56.0} & {88.4} & {64.0} & \underline{65.6} & 67.4\\
            % \rowcolor{Thistle!20}
            \rowcolor{gray!40}
            & {\hspace{1em}}Ours &&\textbf{93.0} & \underline{68.7}& 44.0&47.0 &\underline{70.4} & 85.9 &\textbf{63.5} &\textbf{89.7}&\textbf{66.2} &63.3&\textbf{69.2}\\

            % ----------------------------------------------------
            \midrule \midrule
            \multirow{12}{*}{\rotatebox{90}{{\textbf{Average}}}} &{\hspace{1em}}Continual-FT    & 25.5 & 81.5 & 59.1 & 53.2 & 64.7 & 51.8 & 63.2 & 64.3 & 69.7 & 31.8 & 49.7 & 55.9 \\
            & {\hspace{1em}}LwF \cite{li2017learning} & 36.3 & 86.9 & 72.0 & 59.0 & 73.7 & 60.0 & 73.6 & {74.8} & 80.0 & 37.3 & 58.1 & 64.7 \\
            & {\hspace{1em}}iCaRL \cite{rebuffi2017icarl} & 35.5 & 89.2 & 72.2 & 60.6 & 68.8 & 70.0 & 78.2 & 62.3 & 81.8 & 41.2 & 62.5 & 65.7 \\
            & {\hspace{1em}}LwF-VR \cite{ding2022don} & 29.6 & 87.7 & 74.4 & 59.5 & 72.4 & 63.6 & 77.0 & 66.7 & 81.2 & 43.7 & 60.7 & 65.1  \\
            & {\hspace{1em}}WiSE-FT \cite{wortsman2022robust} & 26.7 & 86.5 & 64.3 & 57.1 & 65.7 & 58.7 & 71.1 & 70.5 & 75.8 & 36.9 & 54.6 & 60.7  \\
            & {\hspace{1em}}ZSCL \cite{zheng2023preventing} & 45.1 & {92.0} & 80.1 & 64.3 & {79.5} & 81.6 & \textbf{89.6} & \underline{75.2} & 88.9 & 64.7 & \textbf{68.0} &75.4 \\
            % & {\hspace{1em}}AweForget \cite{zheng2023preventing} & 46.6 & 93.4 & 82.0 & 67.4 & 82.6 & 83.7 & {90.0} & 75.7 & {89.9} & 66.8 & 68.4 & \\
            \cmidrule{2-14}

            & {\hspace{1em}}L2P \cite{wang2022learning} & 38.0 & 85.2 & 78.2 & 61.3 & 72.9 & 74.9 & {79.7} & 59.1 & {82.0} & 59.7 & 55.4 & 67.9 \\
            & {\hspace{1em}}DualPrompt \cite{wang2022dualprompt} & 37.8 & 84.3 & 78.6 & 60.1 & 71.1 & 73.2 & {79.1} & 73.9 & {82.3} & 55.1 & 52.8 & 68.0 \\
            & {\hspace{1em}}S-Prompts \cite{wang2022s} & 37.5 & 92.5 & 77.5 & 58.2 & 76.4 & 74.1 & {78.8} & 57.9 & {83.0} & 60.8 & 54.4 & 68.3 \\
            & {\hspace{1em}}MoE-Adapter \cite{yu2024boosting} & \textbf{50.2} & 91.9 & \underline{83.1} & \textbf{69.4} & \underline{78.9} & \underline{84.0} & \underline{89.1} & 73.7 & {89.3} & \underline{67.7} & \underline{66.9} & \underline{76.7} \\
            & {\hspace{1em}}DIKI \cite{tang2024mind} & 45.4 & \underline{95.7} & 83.0 & 65.0 & 78.2 & 82.5 & {87.1} & 71.7 & \underline{90.0} & 67.2 & {66.6} & 75.7 \\
            % \rowcolor{Thistle!20}
            % \rowcolor{Thistle!20}
            \rowcolor{gray!40}
            & {\hspace{1em}}Ours &\underline{45.9}&\textbf{95.8}&\textbf{83.3}&\underline{66.5} & \textbf{79.5} &\textbf{84.8} & 87.5 &\textbf{76.6}&\textbf{91.0} & \textbf{69.2} &64.5 &\textbf{76.8} \\
            \midrule\midrule

            % ----------------------------------------------------
            \multirow{12}{*}{\rotatebox{90}{{\textbf{Last}}}} &{\hspace{1em}}Continual-FT & 31.0 & 89.3 & 65.8 & 67.3 & 88.9 & 71.1 & 85.6 & {\textbf{99.6}} & 92.9 & 77.3 & 81.1 & 77.3 \\
            & {\hspace{1em}}LwF \cite{li2017learning} & 26.3 & 87.5 & 71.9 & 66.6 & 79.9 & 66.9 & 83.8 & {\textbf{99.6}} & 92.1 & 66.1 & 80.4 & 74.6 \\
            & {\hspace{1em}}iCaRL \cite{rebuffi2017icarl} & 35.8 & {93.0} & 77.0 & 70.2 & 83.3 & 88.5 & {90.4} & 86.7 & {93.2} & 81.2 & \textbf{81.9} & 80.1 \\
            & {\hspace{1em}}LwF-VR \cite{ding2022don} & 20.5 & 89.8 & 72.3 & 67.6 & 85.5 & 73.8 & 85.7 & {\textbf{99.6}} & 93.1 & 73.3 & 80.9 & 76.6  \\
            & {\hspace{1em}}WiSE-FT \cite{wortsman2022robust} & 27.2 & 90.8 & 68.0 & 68.9 & 86.9 & 74.0 & 87.6 & {\textbf{99.6}} & 92.6 & 77.8 & \underline{81.3} & 77.7  \\
            & {\hspace{1em}}ZSCL \cite{zheng2023preventing} & 40.6 & 92.2 & 81.3 & 70.5 & 94.8 & 90.5 & {91.9} & 98.7 & {93.9} & \textbf{85.3} & 80.2 & 83.6 \\
            % & {\hspace{1em}}AweForget \cite{zheng2023preventing} & 42.4 & 92.7 & 83.2 & 73.2 & 97.0 & 91.8 & {92.2} & 99.1 & {93.9} & 87.4 & 82.6 & \\
            \cmidrule{2-14}

            & {\hspace{1em}}L2P \cite{wang2022learning} & 38.0 & 87.1 & 84.2 & 72.9 & 86.0 & 96.1 & {89.2} & 99.0 & {94.1} & 79.6 & 76.0 & 82.0 \\
            & {\hspace{1em}}DualPrompt \cite{wang2022dualprompt} & 37.8 & 87.1 & 84.6 & 71.8 & 89.2 & 96.3 & {89.1} & 99.1 & \underline{94.5} & 79.9 & 76.5 & 82.3 \\
            & {\hspace{1em}}S-Prompts \cite{wang2022s} & 37.5 & 95.1 & 83.7 & 70.2 & 97.5 & 96.5 & {89.0} & 99.1 & {94.0} & 79.5 & 75.8 & 83.4 \\
            % \rowcolor{Thistle!20}
            & {\hspace{1em}}MoE-Adapter\cite{yu2024boosting} & \textbf{49.8} & 92.2 & 86.1 & \textbf{78.1} & 95.7 & 94.3 & \underline{89.5} & 98.1 & {89.9} & 81.6 & 80.0 & 85.0 \\
            & {\hspace{1em}}DIKI \cite{tang2024mind} & 45.4 & \underline{95.9} & \underline{86.0} & 73.0 & \underline{97.8} & \underline{96.8} & {89.3} &{99.3} & {94.4} & 81.8 & 76.4 & \underline{85.1} \\
            \rowcolor{gray!40}
            & {\hspace{1em}}Ours& \underline{46.8}& \textbf{96.1} & \textbf{86.7} &\underline{75.2} &\textbf{98.1} &\textbf{97.0} &\textbf{89.6} &\underline{99.4}&\textbf{94.7}&\underline{82.8} & 76.7 &\textbf{85.7}\\
        \bottomrule
    \end{tabular}}
    \label{tab:main_result}
    }
\end{table*}
Furthermore, for prompts in the text encoder, we follow ~\cite{tang2024mind} and leverage a batch-wise confidence score to align with the vision encoder, which can be denoted as follow:
\begin{equation}
{E}_{txt} = \frac{1}{B}\sum_{b=1}^{B} \sigma E'_{b},
\end{equation}
where $B$ is the batchsize, and $E'_b$ is the confidence score of images in a batch. 

By integrating the IA-GP strategy with IA-CDDP, VLMs can incrementally acquire knowledge from a stream of tasks. When the distribution of a task is identified as ID, the model assigns prompts to the relevant layers and applies the corresponding weights. This approach mitigates backward forgetting of previously encountered distributions. Conversely, when the distribution is classified as OOD, the model refrains from utilizing prompts, thereby preserving the initial output of the model. This leverages the zero-shot generalization ability of the pre-trained model, effectively reducing forward forgetting for unseen tasks.

\begin{table*}
    \centering
    \caption{Comparison with SOTA on MDCIL benchmark in terms of ``Transfer'', ``Average'', and ``Last'' metrics (\%). ``Ours" denotes our method. We label the best and second methods with \textbf{bold} and \underline{underline} styles. The presented results are derived from the Order-$\text{II}$.}
    {\fontsize{9pt}{11pt}\selectfont
    \renewcommand{\arraystretch}{1.0}
    \resizebox{1.0\linewidth}{!}{
    \begin{tabular}{c>{\raggedright\arraybackslash}p{3cm} >{\centering\arraybackslash}p{1cm} >{\centering\arraybackslash}p{1cm}>{\centering\arraybackslash}p{1cm} >{\centering\arraybackslash}p{1cm} >{\centering\arraybackslash}p{1cm}>{\centering\arraybackslash}p{1cm} >{\centering\arraybackslash}p{1cm} >{\centering\arraybackslash}p{1cm}>{\centering\arraybackslash}p{1cm} >{\centering\arraybackslash}p{1cm} >{\centering\arraybackslash}p{1cm} >{\centering\arraybackslash}p{1.5cm}}
        \toprule
           & {\hspace{1em}} Method & \makecell[c]{\rotatebox{90}{StanfordCars~\cite{krause20133d}}} & \makecell[c]{\rotatebox{90}{Food~\cite{bossard2014food}}} & \makecell[c]{\rotatebox{90}{MNIST~\cite{deng2012mnist}}} & \makecell[c]{\rotatebox{90}{OxfordPet~\cite{parkhi2012cats}}} & \makecell[c]{\rotatebox{90}{Flowers~\cite{nilsback2008automated}}} & \makecell[c]{\rotatebox{90}{SUN397~\cite{xiao2010sun}}} & \makecell[c]{\rotatebox{90}{Aircraft~\cite{maji2013fine}}} & \makecell[c]{\rotatebox{90}{Caltech101~\cite{fei2004learning}}} & \makecell[c]{\rotatebox{90}{DTD~\cite{cimpoi2014describing}}} & \makecell[c]{\rotatebox{90}{EuroSAT~\cite{helber2019eurosat}}} & \makecell[c]{\rotatebox{90}{CIFAR100~\cite{krizhevsky2009learning}}} & \makecell[c]{{\textit{Average}}} \\
  
        \midrule
        
\multirow{2}{*}{\rotatebox{90}{CLIP}}& {\hspace{1em}}Zero-shot & 64.7 & 88.5 & 59.4 & 89.0 & 71.0 & 65.2 & 24.3 & 88.4 & 44.6 & 54.9 & 68.2 & 65.3  \\
& {\hspace{1em}}Full Fine-tune & 89.6 & 92.7 & 99.6 & 94.7 & 97.5 & 81.8 & 62.0 & 95.1 & 79.5 & 98.9 & 89.2 & 89.2  \\
\midrule\midrule
\multirow{12}{*}{\rotatebox{90}{\textbf{Transfer}}} &{\hspace{1em}}Continual-FT & & \textbf{89.5} & \textbf{59.6} & 57.9 & 40.0 & 46.7 & 11.1 & 70.0 & 30.5 & 26.6 & 37.7 & 46.6 \\
& {\hspace{1em}}LwF \cite{li2017learning} & & 87.8 & 58.5 & 71.9 & 46.6 & 57.3 & 12.8 & 81.4 & 34.5 & 34.5 & 46.8 & 53.2 \\
& {\hspace{1em}}iCaRL \cite{rebuffi2017icarl} & & 86.1 & 51.8 & 67.6 & 50.4 & 57.9 & 11.0 & 72.3 & 31.2 & 32.7 & 48.1 & 50.9 \\
& {\hspace{1em}}LwF-VR \cite{ding2022don} & & 88.2 & 57.0 & 71.4 & 50.0 & 58.0 & 13.0 & 82.0 & 34.4 & 29.3 & 47.6 & 53.1 \\
& {\hspace{1em}}WiSE-FT \cite{wortsman2022robust} & & 87.2 & 57.6 & 67.0 & 45.1 & 54.0 & 12.9 & 78.6 & 35.5 & 28.4 & 44.3 & 51.1 \\
& {\hspace{1em}}ZSCL \cite{zheng2023preventing} & & 88.3 & 57.5 & 84.7 & 68.1 & \textbf{64.8} & 21.1 & 88.2 & 45.3 & \textbf{55.2} & \underline{68.2} & 64.1 \\

            \cmidrule{2-14}
& {\hspace{1em}}L2P \cite{wang2022learning} & & 70.6 & 30.7 & 78.3 & 42.8 & 38.3 & 17.4 & 75.3 & 27.4 & 23.1 & 20.7 & 42.5 \\
& {\hspace{1em}}DualPrompt \cite{wang2022dualprompt} & & 79.9 & 46.9 & 85.2 & 51.3 & 45.1 & 9.3 & 82.7 & 29.9 & 42.9 & 47.2 & 52.1 \\
& {\hspace{1em}}S-Prompts \cite{wang2022s} & & 59.8 & 46.2 & 67.7 & 47.5 & 43.8 & 13.5 & 76.8 & 31.4 & 22.6 & 43.5 & 45.3 \\
& {\hspace{1em}}MoE-Adapter \cite{yu2024boosting} & & \underline{88.8} & \underline{59.5} & \underline{89.1} & 69.9 & \underline{64.4} & 18.1 & 86.9 & 43.7 & \underline{54.6} & \underline{68.2} & \underline{64.3} \\
& {\hspace{1em}}DIKI \cite{tang2024mind} & & 85.8 & 55.3 & \textbf{89.5} & \underline{71.1} & 62.9 & \underline{23.7} & \underline{93.6} & 42.1 & 43.4 & 67.9 & 63.5 \\
% \rowcolor{Thistle!20}
\rowcolor{gray!40}
& {\hspace{1em}}Ours & & 85.7 & 59.4 & \underline{89.1} & \textbf{71.3} & 62.7 &\textbf{ 24.4} & \textbf{94.0} &\textbf{ 43.8} & 49.0 & \textbf{68.6} & \textbf{64.9} \\
            
            % ----------------------------------------------------
            \midrule \midrule
\multirow{12}{*}{\rotatebox{90}{{\textbf{Average}}}} &{\hspace{1em}}Continual-FT & 42.1 & 70.5 & \underline{92.2} & 80.1 & 54.5 & 59.1 & 19.8 & 78.3 & 41.0 & 38.1 & 42.3 & 56.2 \\
& {\hspace{1em}}LwF \cite{li2017learning} & 49.0 & 77.0 & {92.1} & 85.9 & 66.5 & 67.2 & 20.9 & 84.7 & 44.6 & 45.5 & 50.5 & 62.2 \\
& {\hspace{1em}}iCaRL \cite{rebuffi2017icarl} & 52.0 & 75.9 & 77.4 & 74.6 & 58.4 & 59.3 & 11.7 & 79.6 & 42.1 & 43.2 & 51.7 & 56.9 \\
& {\hspace{1em}}LwF-VR \cite{ding2022don} & 44.9 & 75.8 & 91.8 & 85.3 & 63.5 & 67.6 & 16.9 & 84.9 & 44.0 & 40.6 & 51.3 & 60.6 \\
& {\hspace{1em}}WiSE-FT \cite{wortsman2022robust} & 52.6 & 79.3 & 91.9 & 83.9 & 63.4 & 65.2 & 23.3 & 83.7 & 45.4 & 40.0 & 48.2 & 61.5 \\
& {\hspace{1em}}ZSCL \cite{zheng2023preventing} & 81.7 & \textbf{91.3} & 91.1 & 91.0 & 82.9 & \textbf{72.5} & 33.6 & 89.7 & \textbf{53.3} & \textbf{62.8} & \underline{69.9} & 74.5 \\
            \cmidrule{2-14}

& {\hspace{1em}}L2P \cite{wang2022learning} & 80.1 & 87.4 & 86.7 & 89.6 & 76.8 & 59.1 & 27.7 & 79.5 & 33.9 & 34.6 & 26.5 & 62.5 \\
& {\hspace{1em}}DualPrompt \cite{wang2022dualprompt} & 78.6 & 88.4 & 89.7 & \underline{91.7} & 80.0 & 62.4 & 23.2 & 85.0 & 41.3 & 51.6 & 50.7 & 67.5 \\
& {\hspace{1em}}S-Prompts \cite{wang2022s} & 79.2 & 86.5 & 89.5 & 87.0 & 78.2 & 61.5 & 25.5 & 83.6 & 41.9 & 36.3 & 47.2 & 65.1 \\
& {\hspace{1em}}MoE-Adapter \cite{yu2024boosting} & \textbf{84.9} & \underline{89.9} & 89.3 & 91.4 & 86.2 & \underline{72.2} & 33.4 & 89.4 & 53.3 & \underline{61.4} & \underline{69.9} & \underline{74.7} \\
& {\hspace{1em}}DIKI \cite{tang2024mind} & 84.8 & 89.0 & 91.3 & \textbf{93.2} & \underline{87.8} & 72.2 & \underline{34.0} & \textbf{94.5} & 50.9 & 53.3 & 69.6 & 74.2 \\
\rowcolor{gray!40}
& {\hspace{1em}}Ours & \underline{82.5} & 89.2 & \textbf{92.3} & \textbf{93.2} & \textbf{88.0} & 70.4 & \textbf{34.3} & \underline{94.4} & \underline{52.4} & 57.9 & \textbf{70.2} & \textbf{75.1} \\
            \midrule\midrule

            % ----------------------------------------------------
\multirow{12}{*}{\rotatebox{90}{{\textbf{Last}}}} &{\hspace{1em}}Continual-FT & 24.0 & 67.3 & 99.1 & 87.4 & 44.3 & 67.0 & 29.5 & 92.3 & 61.3 & 81.0 & \textbf{88.1} & 67.4 \\
& {\hspace{1em}}LwF \cite{li2017learning} & 34.6 & 69.6 & 99.3 & 88.7 & 61.1 & 72.5 & 32.5 & 88.1 & 65.6 & 90.9 & \underline{87.9} & 71.9 \\
& {\hspace{1em}}iCaRL \cite{rebuffi2017icarl} & 46.0 & 81.5 & 91.3 & 82.8 & 66.5 & 72.2 & 16.3 & 91.6 & 68.1 & 83.2 & 87.8 & 71.6 \\
& {\hspace{1em}}LwF-VR \cite{ding2022don} & 27.4 & 61.2 & \underline{99.4} & 86.3 & 60.6 & 70.7 & 23.4 & 88.0 & 61.3 & 84.3 & \textbf{88.1 }& 68.2 \\
& {\hspace{1em}}WiSE-FT \cite{wortsman2022robust} & 35.6 & 76.9 & \textbf{99.5} & 89.1 & 62.1 & 71.8 & 27.8 & 90.8 & 67.0 & 85.6 & 87.6 & 72.2 \\
& {\hspace{1em}}ZSCL \cite{zheng2023preventing} & 78.2 & \textbf{91.1} & 97.6 & 92.5 & 87.4 & \textbf{78.2} & 45.0 & 92.3 & 72.7 & 96.2 & 86.3 & 83.4 \\
\cmidrule{2-14}

& {\hspace{1em}}L2P \cite{wang2022learning} & 80.1 & 89.1 & 99.1 & 93.8 & 96.2 & 76.5 & 40.1 & 86.9 & 73.5 & 86.3 & 84.2 & 82.3 \\
& {\hspace{1em}}DualPrompt \cite{wang2022dualprompt} & 78.6 & 88.3 & 99.2 & 94.1 & 96.5 & 76.8 & 39.8 & 89.0 & 71.6 & 90.7 & 84.9 & 82.8 \\
& {\hspace{1em}}S-Prompts \cite{wang2022s} & 79.2 & 88.1 & 99.1 & 94.3 & 95.8 & 76.3 & 39.9 & 95.5 & 70.1 & \underline{97.6} & 84.4 & 83.8 \\
% \rowcolor{Thistle!20}
& {\hspace{1em}}MoE-Adapter \cite{yu2024boosting} & \textbf{84.1} & 88.5 & 94.0 & 91.8 & 94.1 & 77.8 & \textbf{50.4} & 93.3 & \textbf{77.1} & 87.7 & 86.6 & 84.1 \\
& {\hspace{1em}}DIKI \cite{tang2024mind} & 81.8 & 88.3 & 99.3 & \underline{94.7} & \underline{97.4} & 76.8 & \underline{46.4} & \underline{96.0} & 74.2 & \textbf{98.0} & 86.0 & \underline{85.4} \\
\rowcolor{gray!40}
& {\hspace{1em}}Ours & \underline{82.5} & \underline{88.6} & \underline{99.4} & \textbf{94.9} & \textbf{97.7} & \underline{76.9} & 46.1 & \textbf{96.1} & \underline{74.7} & \textbf{98.0} & 86.6 & \textbf{85.9} \\
        \bottomrule
    \end{tabular}}
    \label{tab:order_ii}
    }
\end{table*}

\begin{table*}
\caption{Comparison with SOTA on 16-shot MTIL benchmark in terms of ``Transfer'', ``Average'', and ``Last'' metrics (\%). ``Ours" denotes our method. We label the best and second methods with \textbf{bold} and \underline{underline} styles.}
    \centering
    \renewcommand{\arraystretch}{1.1}
    \resizebox{1.0\linewidth}{!}{
    \begin{tabular}{c>{\raggedright\arraybackslash}p{3cm} >{\centering\arraybackslash}p{1cm} >{\centering\arraybackslash}p{1cm} >{\centering\arraybackslash}p{1cm} >{\centering\arraybackslash}p{1cm}>{\centering\arraybackslash}p{1cm}>{\centering\arraybackslash}p{1cm} >{\centering\arraybackslash}p{1cm} >{\centering\arraybackslash}p{1cm} >{\centering\arraybackslash}p{1.5cm}}
        \toprule
           & {\hspace{1em}} Method & \makecell[c]{\rotatebox{90}{Aircraft~\cite{maji2013fine}}} & \makecell[c]{\rotatebox{90}{Caltech101~\cite{fei2004learning}}} & \makecell[c]{\rotatebox{90}{CIFAR100~\cite{krizhevsky2009learning}}} & \makecell[c]{\rotatebox{90}{DTD~\cite{cimpoi2014describing}}} & \makecell[c]{\rotatebox{90}{Flowers~\cite{nilsback2008automated}}} & \makecell[c]{\rotatebox{90}{Food~\cite{bossard2014food}}} & \makecell[c]{\rotatebox{90}{StanfordCars~\cite{krause20133d}}} & \makecell[c]{\rotatebox{90}{SUN397~\cite{xiao2010sun}}} & \makecell[c]{{\textit{Average}}} \\
  
        \midrule
        
            \multirow{2}{*}{\rotatebox{90}{CLIP}}
            & {\hspace{1em}}Zero-shot  & 24.8 & \underline{92.9} & 68.4 & 43.8 & \textbf{71.4} & 85.8 & \underline{65.8} & 62.6& 64.4  \\
            & {\hspace{1em}}Full Fine-tune & 62.0 & 96.2 & 89.6 & 79.5 & 97.5 & 92.7 & 89.6 & 81.8  & 86.1\\
            \midrule\midrule
            \multirow{6}{*}{\rotatebox{90}{\textbf{Transfer}}} 
            & {\hspace{1em}}ZSCL \cite{zheng2023preventing}& & 87.3 & 67.7 & 45.4 & 67.8 & \textbf{86.6} & 59.7 & 63.4& 68.3\\

            \cmidrule{2-11}
            & {\hspace{1em}}L2P \cite{wang2022learning} & & 66.7 & 54.3 & 30.6 & 47.3 & 71.5 & 54.6 & 52.4 & 53.9 \\
            & {\hspace{1em}}DualPrompt \cite{wang2022dualprompt} & & 78.8 & 64.4 & 32.0 & 51.7 & 77.5 & 49.4 & 51.3 & 57.9 \\
            & {\hspace{1em}}S-Prompts \cite{wang2022s}& & 70.3 & 52.7 & 31.5 & 54.8 & 74.0 & 55.4 & 50.0 & 55.5 \\
            & {\hspace{1em}}DIKI \cite{tang2024mind} & & 92.7 & \underline{68.8} & 44.1 & \underline{70.0} & \underline{86.2} & 65.1 & \textbf{65.5} & \underline{70.3} \\
            \rowcolor{gray!40}
            & {\hspace{1em}}Ours &&\textbf{93.2} &\textbf{68.9} & \underline{44.5} &\textbf{71.4} &85.5&\textbf{66.1}&\underline{65.0}&\textbf{70.9}\\

            % ----------------------------------------------------
            \midrule \midrule
            \multirow{6}{*}{\rotatebox{90}{{\textbf{Average}}}} 
            & {\hspace{1em}}ZSCL \cite{zheng2023preventing}&33.5& 90.5& 74.7&\underline{58.5} &79.7 &87.7 & 64.8 &64.8 &69.3\\
            \cmidrule{2-11}

            & {\hspace{1em}}L2P \cite{wang2022learning} & 30.2 & 84.5 & 70.1 & 51.9 & 69.6 & 77.1 & 60.0 & 55.2 & 62.3 \\
            & {\hspace{1em}}DualPrompt \cite{wang2022dualprompt} & 36.5 & 89.5 & 72.5 & 52.7 & 72.3 & 80.8 & 56.1 & 54.2 & 64.3 \\
            & {\hspace{1em}}S-Prompts \cite{wang2022s} & 30.6 & 86.8 & 70.0 & 51.7 & 74.3 & 78.5 & 60.7 & 53.0 & 63.2 \\
            & {\hspace{1em}}DIKI \cite{tang2024mind}& \underline{41.3} & \textbf{95.3} & \underline{76.5} & \underline{58.5} & \underline{82.2} & \underline{86.4} & \underline{68.2} & \textbf{66.6} & \underline{71.9} \\
            \rowcolor{gray!40}
            & {\hspace{1em}}Ours &\textbf{42.5}& \underline{94.8} &\textbf{77.6} &\textbf{59.3}&\textbf{82.5}&\textbf{86.5} &\textbf{69.6}&\underline{65.4}&\textbf{72.5} \\
            \midrule\midrule

            % ----------------------------------------------------
            \multirow{6}{*}{\rotatebox{90}{{\textbf{Last}}}} 
            & {\hspace{1em}}ZSCL \cite{zheng2023preventing}& 27.7& 90.9 & 74.4 & 64.7 & 90.2 & \textbf{89.2} & \textbf{80.6} &\underline{74.6}&74.0 \\
            \cmidrule{2-11}

            & {\hspace{1em}}L2P \cite{wang2022learning}& 30.2 & 87.1 & 75.4 & 64.7 & 91.9 & 86.4 & 76.1 & 74.7 & 73.3 \\
            & {\hspace{1em}}DualPrompt \cite{wang2022dualprompt}& 36.5 & 91.0 & 75.1 & 65.1 & 92.9 & 86.2 & 76.2 & 74.2 & 74.7 \\
            & {\hspace{1em}}S-Prompts \cite{wang2022s}  & 30.6 & 89.2 & 75.8 & 63.8 & 93.9 & 86.2 & 76.7 & 73.9 & 73.8 \\
            & {\hspace{1em}}DIKI \cite{tang2024mind} & \underline{41.3} & \underline{95.6} & \textbf{79.0} & \underline{67.3} & \underline{94.4} & 86.8 & 77.6 & 74.4 & \underline{77.1} \\
            \rowcolor{gray!40}
            & {\hspace{1em}}Ours&\textbf{42.5}&\textbf{95.8}&\underline{78.6} &\textbf{68.1} &\textbf{95.3}&\underline{87.5}&\underline{79.2}& \textbf{74.8} &\textbf{77.7}\\
        \bottomrule
    \end{tabular}}
    \label{tab:few-shot}
\end{table*}

\section{Experiments}
\subsection{Experimental Settings}
\subsubsection{Dataset} 
We evaluate our approach in the MTIL benchmark, which includes 11 diverse datasets: Aircraft~\cite{maji2013fine}, Caltech101~\cite{fei2004learning}, CIFAR100~\cite{krizhevsky2009learning}, DTD~\cite{cimpoi2014describing}, EuroSAT~\cite{helber2019eurosat}, Flowers~\cite{nilsback2008automated}, Food~\cite{bossard2014food}, MNIST~\cite{deng2012mnist}, OxfordPet~\cite{parkhi2012cats}, StanfordCars~\cite{krause20133d} and SUN397~\cite{xiao2010sun}. These datasets collectively encompass 1201 classes, each characterized by distinct distributions. The details of datasets are provided in the supplemental materials. 

\subsubsection{Metrics} 
We evaluate our method using three metrics proposed by ~\cite{zheng2023preventing}: ``Transfer'', ``Last'', and ``Avg''. The ``Transfer" metric, unconventional in traditional incremental learning approaches, quantifies the forward forgetting of the model's zero-shot generalization ability across tasks. Specifically, for a given task $i$, it is computed as the average performance across all unseen tasks $i+1, i+2,\cdots,\mathcal{T}$. The ``Last" metric assesses the model's ability to acquire knowledge from new tasks while minimizing catastrophic forgetting of previously encountered tasks, addressing backward forgetting. For task $i$, it is calculated by averaging performance over all seen tasks $i, i-1,\cdots,1$. The ``Avg" metric provides a holistic evaluation by integrating both backward and forward forgetting, computed as the mean performance across all tasks $\mathcal{T}$ at each step of the incremental learning process. These metrics are designed to assess the performance of incremental learning methods, collectively offering a robust evaluation framework. 

\begin{table}[t]
  \centering
  \caption{Comparison of the use of memory buffers and the scale of trainable parameters. $\dagger$ represents ours with 8-layer IA-GP strategy.}
  \label{tab:mem_para}
  {\fontsize{9pt}{11pt}\selectfont
  \renewcommand{\arraystretch}{0.99}
  \begin{tabular}{l|>{\centering\arraybackslash}p{0.85cm}|>{\centering\arraybackslash}p{0.85cm}|>{\centering\arraybackslash}p{0.85cm}|>{\centering\arraybackslash}p{0.85cm}|>{\centering\arraybackslash}p{0.85cm}}
    \toprule
    Method & Mem. & Param. & Trans. & Avg. & Lst. \\
    \midrule
    iCaRL \cite{rebuffi2017icarl} & \checkmark & 211M & 50.4 & 65.7 & 80.1 \\
    ZSCL \cite{zheng2023preventing}& \checkmark & 211M & 68.1 & 75.4 & 83.6 \\
    MoE \cite{yu2024boosting}& \textbf{$\times$} & 59.8M & 68.9 & 76.7 & 85.0 \\
    DIKI \cite{tang2024mind}& \textbf{$\times$} & \textbf{1.8M} & 67.4 & 75.7 & 85.1 \\
    \rowcolor{gray!40}
    Ours$\dagger$ & \textbf{$\times$} & \textbf{1.8M} & 68.7 & 76.4 & 85.4 \\
    \rowcolor{gray!40}
    Ours & \textbf{$\times$} & 2.4M & \textbf{69.2} & \textbf{76.8} & \textbf{85.7} \\
    \bottomrule
  \end{tabular}
  }
\end{table}
\subsubsection{Comparison Methods} 
We compare our method with two categories of methods in incremental learning: 1) Full-parameter fine-tuning methods, which typically leverage knowledge distillation or rehearsal-based techniques and update all parameters: Continual-FT, LwF~\cite{li2017learning}, iCaRL~\cite{rebuffi2017icarl}, LwF-VR~\cite{ding2022don}, WiSE-FT~\cite{wortsman2022robust}, and ZSCL~\cite{zheng2023preventing}. 2) PEFT methods, which adapt models to new distributions by maintaining only a small set of parameters: L2P~\cite{wang2022learning}, DualPrompt~\cite{wang2022dualprompt}, S-Prompt~\cite{wang2022s}, MoE-Adapter~\cite{yu2024boosting} and DIKI~\cite{tang2024mind}. Our proposed method also falls within PEFT methods.

\begin{table}[t]
  \centering
  \caption{The ablation experiments for each strategy of our IAP method. Asterisk (*) denotes the original performance reported in the DIKI paper.}
  \label{tab:ablation}
  {\fontsize{9pt}{11pt}\selectfont
  \renewcommand{\arraystretch}{1.12}
  \begin{tabular}{l|>{\centering\arraybackslash}p{1.1cm}|>{\centering\arraybackslash}p{1.1cm}|>{\centering\arraybackslash}p{1.1cm}}
    \toprule
    Method & Tran. & Avg. & Lst. \\
    \midrule
    DIKI & 67.4 & 75.7 & 85.1 \\
    DIKI* & 68.7 & 76.3 & 85.1 \\
    Ours w/o IA-CDDP & 68.6 & 76.6 & 85.5 \\
    Ours w/o IA-GP & 69.1 & 76.5 & 85.3 \\
    \rowcolor{gray!40}
    Ours & \textbf{69.2} & \textbf{76.8} & \textbf{85.7} \\
    \bottomrule
  \end{tabular}
  }
\end{table}
\subsubsection{Implementation Details} 
To ensure fair comparisons, we maintain the same backbone as ~\cite{zheng2023preventing}, utilizing CLIP ViT-B/16 as our vision-language model. We employ Stochastic Gradient Descent (SGD) as the optimizer and the model is trained for 10 epochs across all tasks. Following ~\cite{tang2024mind}, we set the learning rate to 5.0 and apply prompts to the first 8 transformer layers of the text encoder, with a prompt length of 8. For the image encoder, we adopt the IA-GP strategy and integrate the prompts into all transformer layers, with the Gumbel sampling temperature set to 3.0. The batch size was set to 128 during training and 256 during inference. Experiments ere conducted using FP16 precision to minimize the computational costs of training and inference. In the first stage of the IA-CDDP strategy, we employ an upper bound of 0.8 and a lower bound of 0.2, a perturbation of $1\times 10^{-6}$ is introduced to all sampled values during Gumbel sampling to mitigate numerical instability. In the second stage, we set $K=5$ to select the top five most relevant classes, we also follow ~\cite{tang2024mind} and apply a regularization term of $10^{-7}$ to the task-wise covariance matrix and $10^{-3}$ to the class-wise covariance matrix to address numerical instability. All experiments are performed on an NVIDIA 4090 GPU.

\begin{table}[t]
  \centering
  \caption{Comparison of our method with two additional adaptive prompting strategies. }
  \label{tab:baselines}
  {\fontsize{9pt}{11pt}\selectfont
  \renewcommand{\arraystretch}{1.0}
  \begin{tabular}
  {l|>{\centering\arraybackslash}p{1.5cm}|>{\centering\arraybackslash}p{1.5cm}|>{\centering\arraybackslash}p{1.5cm}}
    \toprule
    Method & Tran. & Avg. & Lst. \\
    \midrule
    Random Layer& 68.8 & 75.9 & 84.5 \\
    Soft Gumbel& 68.6 & 76.3 & 85.3 \\
    \rowcolor{gray!40}
    Ours & \textbf{69.2} & \textbf{76.8} & \textbf{85.7} \\
    \bottomrule
  \end{tabular}
  }
\end{table}
\subsection{Experimental Results} 
\subsubsection{Main Results} 
We follow ~\cite{zheng2023preventing} and conduct our main experiment through Order-$\text{I}$ task sequence, where the model is trained incrementally on a series of tasks in the following order: Aircraft~\cite{maji2013fine}, Caltech101~\cite{fei2004learning}, CIFAR100~\cite{krizhevsky2009learning}, DTD~\cite{cimpoi2014describing}, EuroSAT~\cite{helber2019eurosat}, Flowers~\cite{nilsback2008automated}, Food~\cite{bossard2014food}, MNIST~\cite{deng2012mnist}, OxfordPet~\cite{parkhi2012cats}, StanfordCars~\cite{krause20133d} and SUN397~\cite{xiao2010sun}. The results are presented in Table \ref{tab:main_result}. We provide a detailed performance comparison of our proposed method against other SOTA methods within the MTIL benchmark. At the top of the table, we report the zero-shot generalization ability of the CLIP model and its performance under full-parameter fine-tuning as baselines. Our proposed method, denoted as ``Ours'' in the table, outperforms all SOTA methods for MTIL across three key metrics. Specifically, our method achieves the highest performance on the ``Transfer'' metric for 4 out of 11 tasks, the ``Average'' metric for 7 tasks, and the ``Last'' metric for 6 tasks. Furthermore, our approach establishes new SOTA performance in terms of the overall mean values across all three metrics. While the MoE-Adapter~\cite{yu2024boosting} method demonstrates the closest performance to ours, our approach achieves this with a significantly smaller number of learnable parameters. More detailed results for each incremental learning session can be found in the supplemental materials.

\subsubsection{Order-$\text{II}$ Result} 
We evaluate our method in the second task order, where the tasks are acquired through the following sequence: StanfordCars~\cite{krause20133d}, Food~\cite{bossard2014food}, MNIST~\cite{deng2012mnist}, OxfordPet~\cite{parkhi2012cats}, Flowers~\cite{nilsback2008automated}, SUN397~\cite{xiao2010sun}, Aircraft~\cite{maji2013fine}, Caltech101~\cite{fei2004learning}, DTD~\cite{cimpoi2014describing}, EuroSAT~\cite{helber2019eurosat} and CIFAR100~\cite{krizhevsky2009learning}. The experimental results are provided in Table \ref{tab:order_ii}. In the second order, our methods also achieve the optimal performance among all the metrics ``Transfer'', ``Average'' and ``Last'', which demonstrates that our method is robust to variations in task order.

\subsubsection{Few-shot Incremental Learning} 
We follow~\cite{tang2024mind} and evaluate our method in the 16-shot MTIL setting, where only 16 samples per task are available during model training. The comparison results are provided in Table~\ref{tab:few-shot}, which uses the same metrics as the main results. Our method also outperforms all other SOTA approaches. This reveals that our method can effectively learn new tasks with limited data while exhibiting minimal forgetting of previously acquired knowledge.

\begin{figure}[t]
  \centering
  \includegraphics[width=0.47\textwidth]{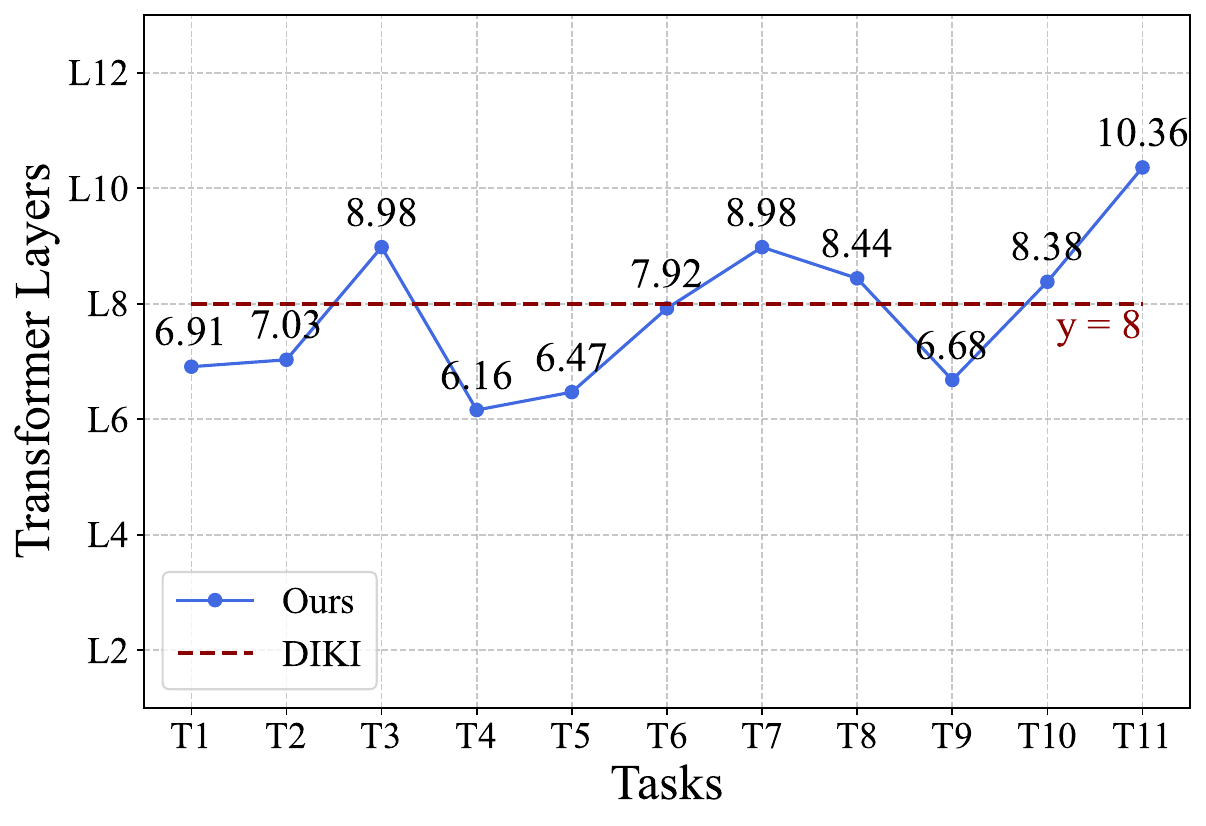}
  \caption{Mean prompting layers for different tasks. The tasks with complex distributions tend to allocate more prompting layers.}
  \label{fig:meanlayers}
\end{figure}

\subsubsection{Trainable Parameters and Memory Buffer} 
We present a comparison of the trainable parameters and the requirements for the rehearsal memory buffer for our method and four other SOTA approaches in Table \ref{tab:mem_para}. Leveraging PEFT strategies, our method eliminates the need for external storage of representative samples or features. Our approach utilizes only $4\%$ of the trainable parameters compared to the MoE-Adapter~\cite{yu2024boosting} method. This substantial reduction is achieved because our method does not require the training of numerous experts; instead, for each task, only the corresponding prompt pool and prompt gate module are trained. Furthermore, when the IA-GP strategy is applied only to the first 8 layers (the same setting as DIKI~\cite{tang2024mind}), our method can achieve higher performance while keeping the trainable parameters at 1.8M (as shown in ``Ours$\dagger$'' in Table~\ref{tab:mem_para}).

\subsubsection{Instance-Aware Gated Baselines}
To provide a more intuitive demonstration of the effectiveness of our IA-CP strategy, we conducted two additional baseline experiments with adaptive prompting strategies: 1) Random prompting layers. For each instance, the model randomly determines the number of prompting layers. Specifically, for every transformer layer, there is a 50\% probability that a prompt will be injected. 2) Soft Gumbel Softmax: rather than enforcing a binary decision (0 or 1) on whether to inject a prompt into a transformer layer, the model outputs a soft value to control the prompting process. We compared these two baselines against the results of our main experiment, as shown in Figure~\ref{tab:baselines}. Under identical experimental settings, our proposed method consistently outperformed both baselines across all the ``Transfer'' ``Average'' and ``Last'' metrics.

\begin{figure*}[t]
    \centering
    \subfloat[fig 5]{\includegraphics[width=0.33\textwidth]{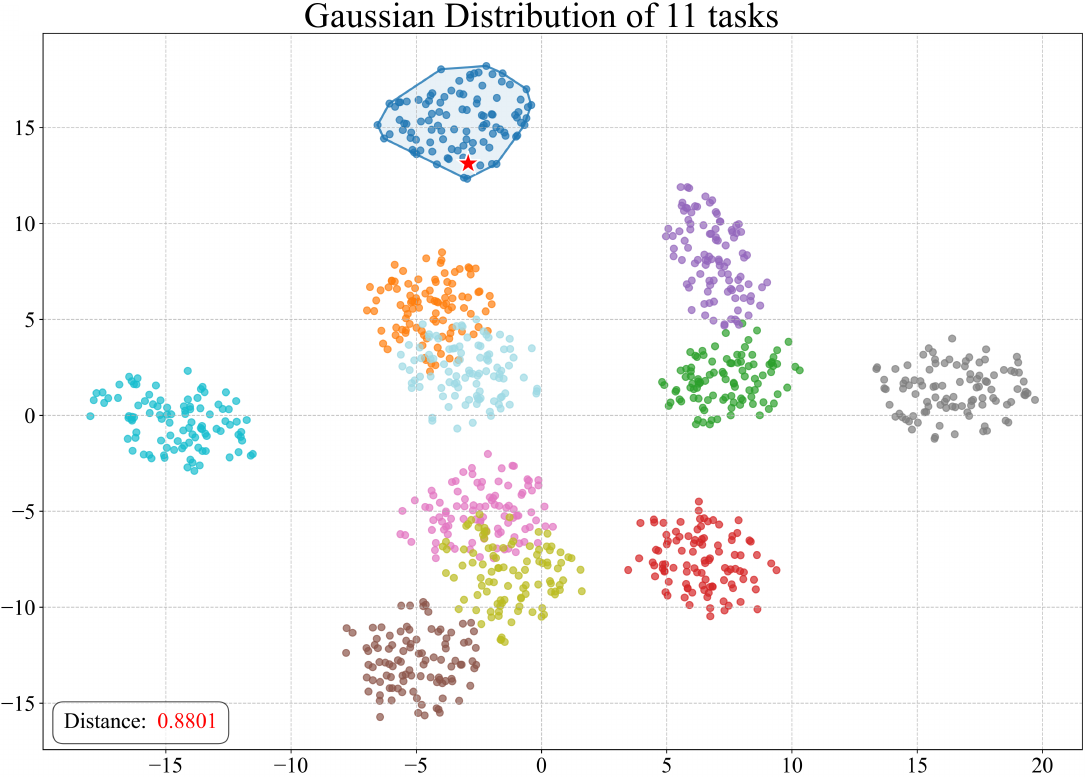}}
    \hfil
    \subfloat[fig 6]{\includegraphics[width=0.33\textwidth]{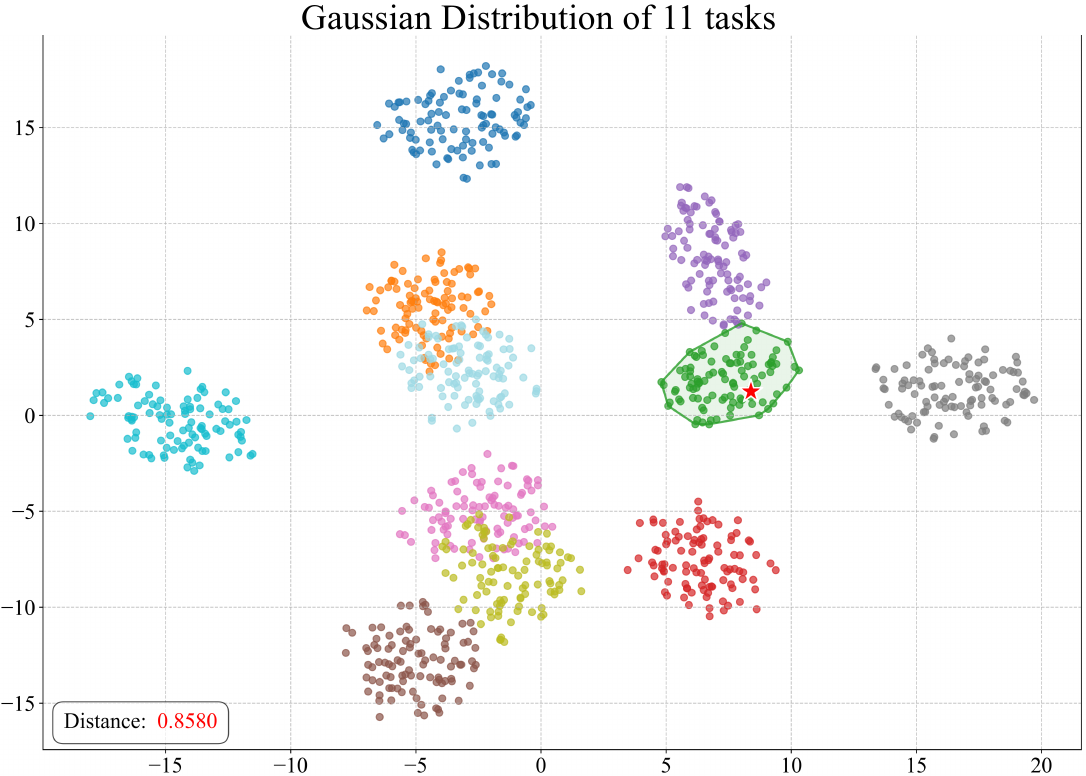}}
    \hfil
    \subfloat[fig 7]{\includegraphics[width=0.33\textwidth]{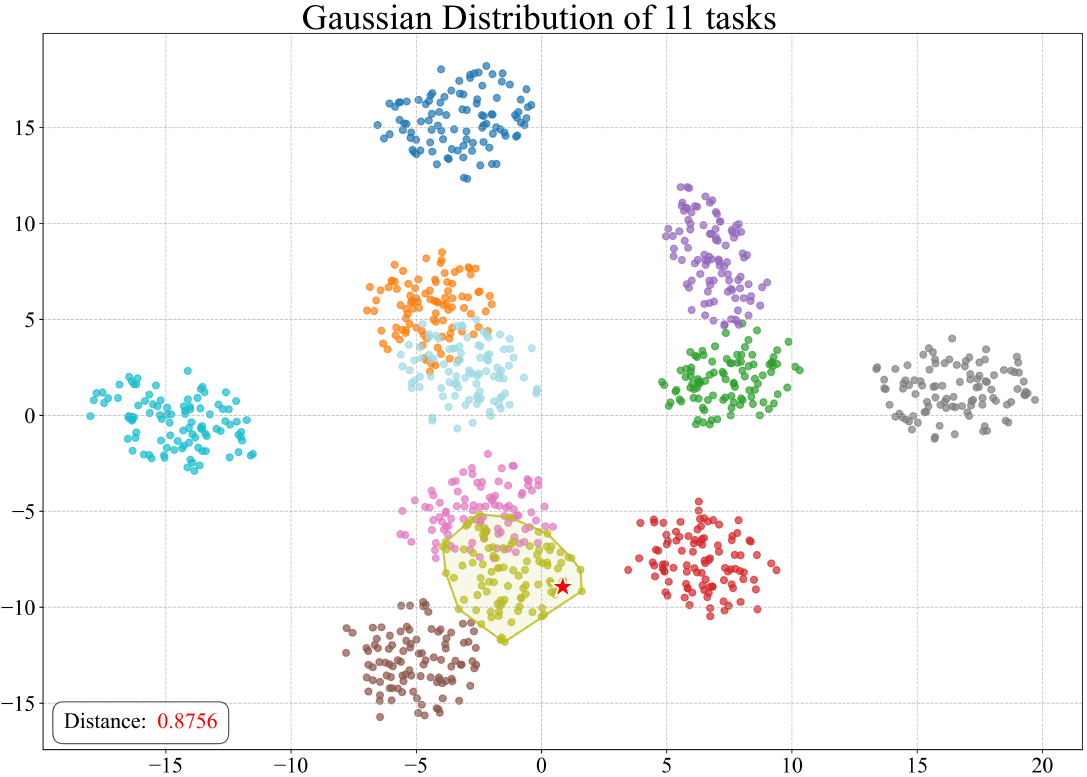}}
    \\ % 换行以创建第二行子图
    \subfloat[fig 1]{\includegraphics[width=0.33\textwidth]{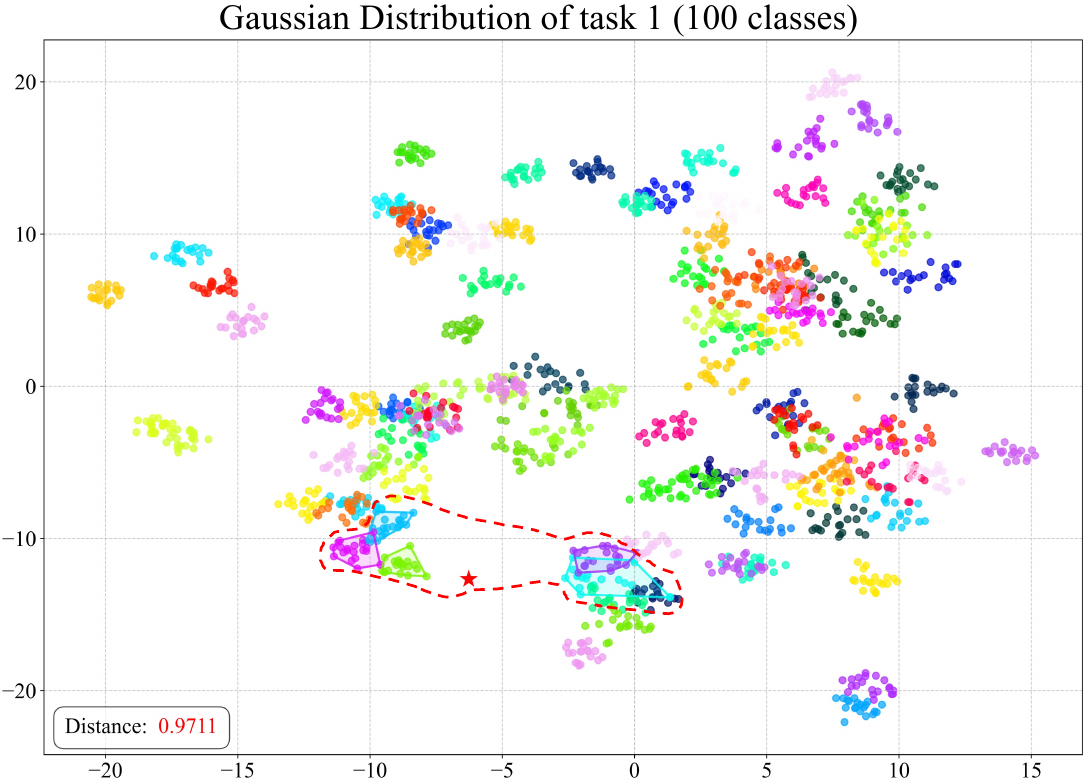}}
    \hfil
    \subfloat[fig 2]{\includegraphics[width=0.33\textwidth]{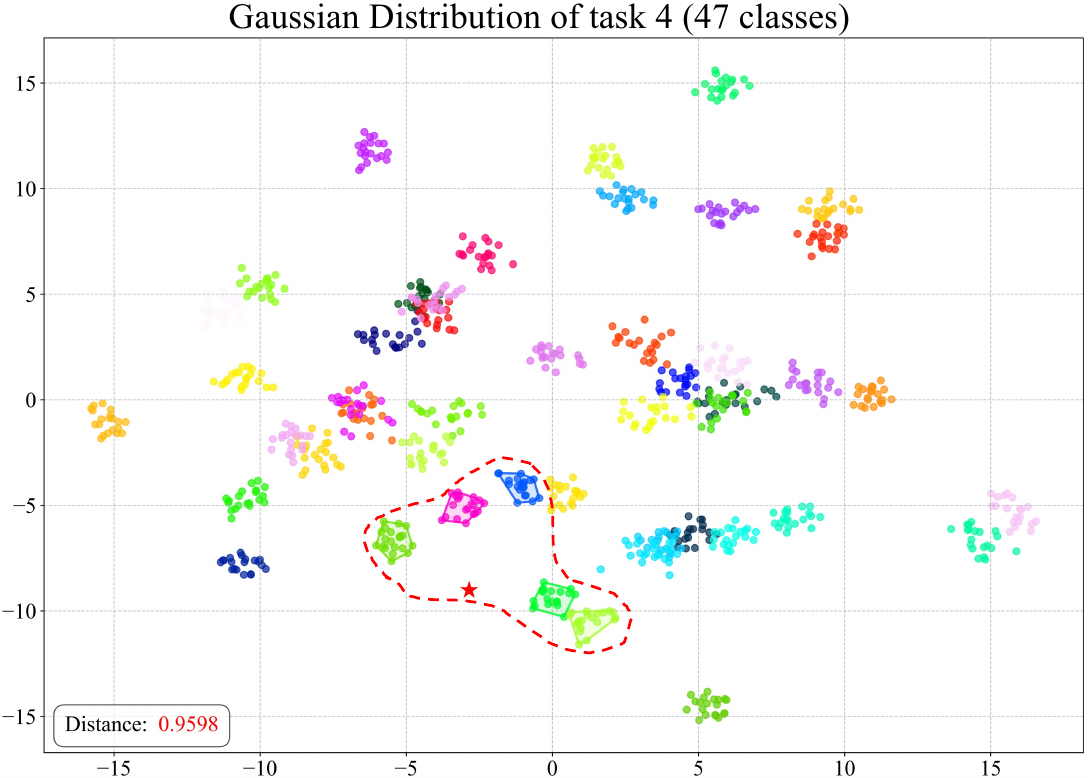}}
    \hfil
    \subfloat[fig 3]{\includegraphics[width=0.33\textwidth]{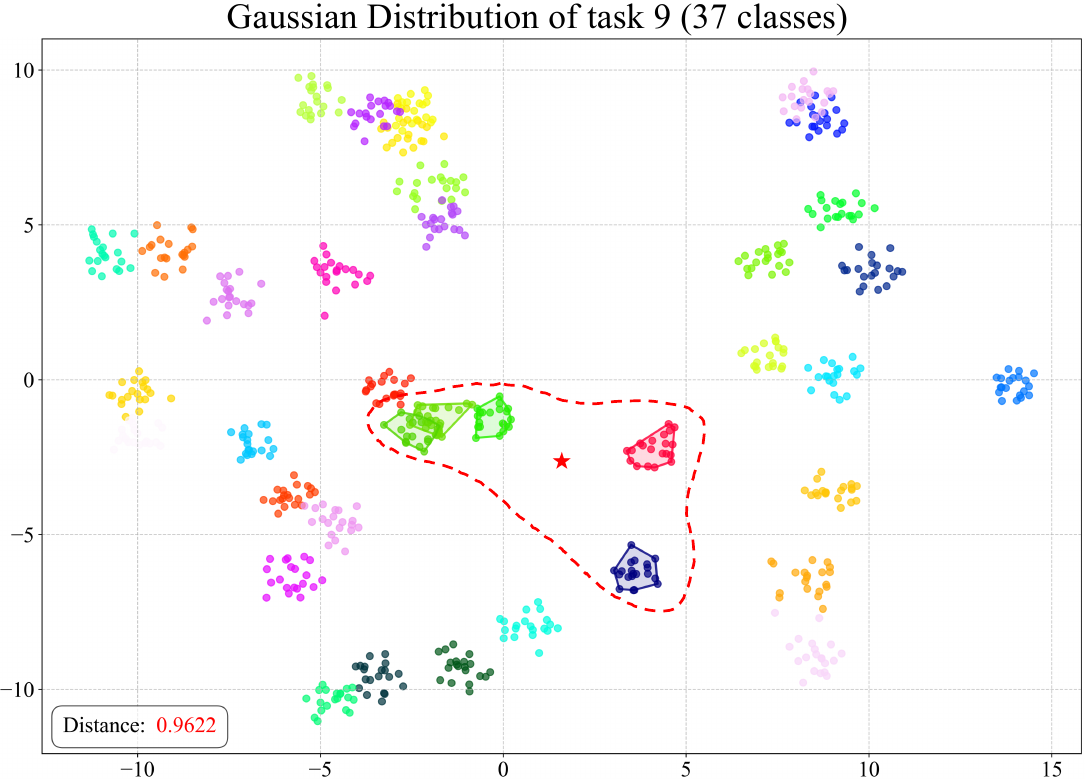}}
    \caption{The classification process of our IA-CDDP module for test samples across tasks 1, 4, and 9. As illustrated in the first row of the figure, the first stage of IA-CDDP constructs distributions at the task level and performs inference based on the similarity between samples and each distribution. Test samples with either excessively low or high confidence scores are directly classified, while samples with intermediate confidence scores proceed to the second stage of IA-CDDP. At this stage, we construct distributions for all classes within the task and calculates similarities with the 5 nearest class distributions, highlighted by the red dashed circle in the second row. The mean of these distribution similarities serves as the confidence score for the test samples, which subsequently guides the following prompting process.}
    \label{fig:distributions}
\end{figure*}

\subsubsection{Effect of Instance-Aware Gated Prompting Strategy} 
We conduct a statistical analysis of the mean prompting layers employed by our IA-GP strategy across different distributions, as illustrated in Figure \ref{fig:meanlayers}. The red line represents the prompt injection strategy employed by DIKI~\cite{tang2024mind}, which consistently prompting the first 8 layers. In contrast, our method, benefiting from the IA-GP strategy, demonstrates greater flexibility by adaptively determining whether to apply prompts to different transformer layers at the instance level. Experimental results indicate that for tasks with more complex distributions, such as SUN397 (397 classes) and Food101 (101 classes), the IA-GP strategy tends to incorporate more prompts to effectively capture intricate distributional information. Conversely, for relatively simpler datasets like DTD (47 classes) and EuroSAT (10 classes), the strategy employs fewer prompts, thereby achieving superior performance while avoiding overfitting. 
%We also provide the specific prompting layers for each task, presented as a heatmap. For details, please refer to Figure~\ref{fig:heatmap} of the supplemental materials.

\subsubsection{Effect of Instance-Aware Class-Distribution-Driven Prompting Strategy} 
In Figure~\ref{fig:distributions}, we visualize the distributions of all 11 tasks and the class distributions of three specific tasks: Aircraft (100 classes), DTD (47 classes), and OxfordPet (37 classes). For each task, we use a corresponding instance as an example, denoted by a star. The reference distance, displayed in the lower left corner of each graph, serves as the confidence score, with values closer to 1 signifying higher confident. The top row of the figure illustrates the distributions of the 11 tasks. The DIKI~\cite{tang2024mind} method leverages the similarity between an instance and the tasks distribution to derive confidence scores. In contrast, our IA-CDDP strategy incorporates a second stage that computes the average of the 5 nearest class distributions (highlighted by red dotted circles) at the instance level. The reference distances demonstrate that our IA-CDDP strategy assigns more appropriate instance-aware confidence scores, thereby enhancing the efficacy of subsequent prompts. We also conduct ablation study on the bounds of IA-CDDP strategy, the details are provided in the supplemental materials.

\subsubsection{Quantitative Analysis}
Our method is implemented on a classic prompt-based MTIL method DIKI~\cite{tang2024mind}. We perform a quantitative analysis of our proposed IA-GP and IA-CDDP strategies and the results are shown in Table \ref{tab:ablation}. For the baseline method, we observed that the reproduced results (DIKI) are lower than the papers' results (DIKI*). Compared to baseline DIKI, both IA-GP and IA-CDDP can achieve a performance improvement.

\section{Conclusions}
In this paper, we propose the Instance-Aware Prompting method to solve the challenges of backward forgetting and forward forgetting faced by pre-trained VLMs during incremental learning of new tasks. Our method comprises two meticulous strategies designed at the instance level. The Instance-Aware Gated Prompting strategy dynamically controls whether to prompt at the instance level by incorporating a prompt gate module for each transformer layer. The Instance-Aware Class-Distribution-Driven Prompting strategy assigns more precise weights to each instance through a two-stage determination process. By integrating these two strategies, our method simultaneously resolves the questions of whether to apply prompt and how much prompt to apply, enabling pre-trained VLMs to adapt more flexibly when incrementally learning new tasks. Extensive experimental results demonstrate that the IAP method represents the current SOTA approach in the MTIL benchmark.
\bibliographystyle{IEEEtran}
\bibliography{IEEEfull}
\end{document}